\documentclass{article}

\PassOptionsToPackage{square,sort,comma,numbers}{natbib}

\usepackage[final]{neurips_2024}
\usepackage{textcomp}
\usepackage[utf8]{inputenc}
\DeclareUnicodeCharacter{202F}{\,}
\usepackage[utf8]{inputenc} % allow utf-8 input
\usepackage[T1]{fontenc}    % use 8-bit T1 fonts
\usepackage[hidelinks]{hyperref}
\usepackage{url}            % simple URL typesetting
\usepackage{booktabs}       % professional-quality tables
\usepackage{amsfonts}       % blackboard math symbols
\usepackage{nicefrac}       % compact symbols for 1/2, etc.
\usepackage{microtype}      % microtypography
\usepackage{xcolor}         % colors
\usepackage{amsmath,amssymb,amsfonts}
\usepackage{mathtools}
\usepackage{stmaryrd}
\usepackage{bbm}
\usepackage{amsmath}
\usepackage{stmaryrd}
\usepackage{tgtermes}
\usepackage{newtxtext}
\usepackage{newtxmath}
\usepackage{bm}
\usepackage{endnotes}

\usepackage{amssymb,amsfonts}
\usepackage{caption}
\usepackage{subcaption}
\usepackage{graphicx}

\usepackage{url}
\usepackage{enumerate}
\usepackage{multicol}
\usepackage{multirow}
\usepackage{mathtools}
\usepackage{tabularray}
\usepackage{cals}
\usepackage{algorithm}
\usepackage{algpseudocode}
\usepackage{tikz}
\usepackage{color}
\usepackage{array}
\usepackage{multirow}

\title{Sliced-Wasserstein-based Anomaly Detection and Open Dataset for Localized Critical Peak Rebates}

\author{
Julien Pallage\thanks{Corresponding author: \texttt{julien.pallage@polymtl.ca}.} \\
Dept. of Electrical Engineering\\
Polytechnique Montréal, GERAD \& Mila \\
\And
Bertrand Scherrer \\
Data Science, Hilo \\
Hydro-Québec\\
\And 
Salma Naccache\\
Data Science, Hilo\\
Hydro-Québec\\
\And
Christophe Bélanger\\
Open Data | Data and Analytics\\
Hydro-Québec\\
\And
Antoine Lesage-Landry\\
Dept. of Electrical Engineering\\
Polytechnique Montréal, GERAD \& Mila\\
}

\begin{document}

\maketitle

\begin{abstract}
  In this work, we present a new unsupervised anomaly (outlier) detection (AD) method using the sliced-Wasserstein metric. This filtering technique is conceptually interesting for MLOps pipelines deploying machine learning models in critical sectors, e.g., energy, as it offers a conservative data selection. Additionally, we open the first dataset showcasing localized critical peak rebate demand response in a northern climate. We demonstrate the capabilities of our method on synthetic datasets as well as standard AD datasets and use it in the making of a first benchmark for our open-source localized critical peak rebate dataset.
\end{abstract}
\section{Introduction}
Québec, Canada, stands as an outlier in the electrical grid decarbonization paradigm. As there is a global tendency to invest in intermittent renewable energy sources to decarbonize grids worldwide, Québec, while not totally stranger to this trend, is generally mostly carbon-free thanks to its impressive hydroelectric power capacity. One of its main challenges comes from its northern climate, its reliance on electric baseboard heating systems for residential heating, and its unrestrictive home insulation policies~\cite{Gerbet_2023}. During glacial winter days, as electric heaters are all running at once, and peak consumption hours hit, Québec's hydropower capacity can be reached~\cite{whitmore2024energie}. To accommodate winter peak power needs, Hydro-Québec, the state-owned company in charge of electric power generation, transmission, and distribution, must operate its only on-grid thermal power plant and import electricity from neighbouring provinces and states. These imports are usually expensive and produce much more greenhouse gas emissions in comparison to local energy sources~\cite{Hydro-Québec_2023}. 

To remediate this issue without solely relying on the deployment of new generation and transmission infrastructures, demand response (DR) initiatives have flourished in the province~\cite{PELLETIER2022107080}. Demand response can be defined as changes in normal electrical power consumption patterns by end-use customers in response to a signal, e.g.,  financial incentives or control setpoints~\cite{albadi2007DR}. With Québec's long tradition of fixed electricity rates, one of its main DR mechanisms is critical peak rebates (CPR). Residential customers enrolled in a CPR program receive financial compensation during pre-specified time periods, referred to as challenges, for reducing their energy consumption with respect to their expected baseload~\cite{mercado2014enabling}. CPR programs are purely voluntary and virtually penalty-free. They thus depend on consumers' goodwill, motivation, and sensitivity. Yet, as we have seen in our work, CPR events can be powerful tools for shifting power consumption before and after each event. 

We work with a variation of CPR, viz., localized CPR (LCPR), in which the events are called for localized relief in the grid instead of being cast for the whole system. LCPR can diversify the types of services offered by typical CPR programs, e.g., they can alleviate stress on local equipment like substation transformers~\cite{feng2024gridedge} or they can be paired with generation forecasts of distributed energy resources (DERs), e.g., roof solar panels and private windfarms, to balance demand and generation during peak hours. LCPR is highly underexplored in the literature and is a valuable application to benchmark trustworthy machine learning (ML) models. Indeed, the higher spatial granularity, the critical aspect of the task, the dependence on behavioural tendencies, and the lower margin for error require the deployment of forecasting models that offer performance guarantees \cite{Venzke2021Worstcase}, robustness to noise~\cite{Chen2020_DRO}, interpretability~\cite{Leilani2018_intepretableML}, physical constraints satisfaction~\cite{Misyris2020PINN_PES}, a sense of prediction confidence~\cite{Jospine2022_handson_BNN, wilson2020bayesian}, or a combination of them~\cite{pallage2024wadiro}. Being able to predict and utilize localized peak-shaving potential in the electrical grid, through programs similar to LCPR, could accelerate the integration of DERs and, specifically for Québec, phase out its dependence on fossil energy-based imports.  To the best of our knowledge, no published open-source datasets are showcasing either LCPR or CPR schemes, in a northern climate. 

As with most smart grid applications, dataset quality can highly influence the performance of ML models when used for training. The adversarial properties stemming from the amalgam of numerical errors, noise in sensor readings, telemetry issues, meter outages, and unusual extreme events can disrupt the prediction quality of ML models. Anomaly detection (AD) and outlier filtering are thus primordial in a reliable ML pipeline~\cite{stiasny2022closing}. Unsupervised AD methods are preferable as they do not need human-made labels and their hyperparameters can be tuned simultaneously with other ML models' included in the loop. Popular unsupervised AD methods~\cite{goldstein2016AD} include local outlier factor (LOF)~\cite{Breuning2000LOF}, isolation forest \cite{Liu2008IsolationForest}, k-nearest neighbours (KNN) \cite{angiulli2002knn}, connectivity-based outlier factor~\cite{Tang2002CLOF}, and one-class support vector machine (SVM)~\cite{Amer2013OneClassSVM}. These methods either use clustering or local density to assign an outlier score. We are interested in optimal transport-based (OT) metrics for AD. Reference~\cite{ducoffe2019anomaly} proposed a Wasserstein generative adversarial network for AD tasks and authors from~\cite{wang2024WOOD} have designed a differentiable training loss function using the Wasserstein metric for deep classifiers. To the best of our knowledge, no unsupervised OT method has been proposed yet for AD.

%\subsection{Contributions}
In this work, we address the lack of open-source datasets showcasing CPR demand response mechanisms in northern climates by releasing two years of aggregated consumption data for customers participating in an LCPR program in Montréal, Québec, Canada\footnote{Available at \url{https://github.com/jupall/lcpr-data}, as well as, \\ \url{https://donnees.hydroquebec.com/explore/dataset/consommation-clients-evenements-pointe}}. These customers are spread in three adjacent distribution substations. With it, we hope to stimulate research on trustworthy ML models for demand response applications. We also address the challenges of training ML models with unfiltered smart grid data by proposing a new simple unsupervised outlier filter leveraging the Sliced-Wasserstein distance~\cite{bonneel2015sliced}. We showcase the performance of this filter on standard anomaly detection datasets and leverage it to present a benchmark for the prediction of the localized energy consumption of LCPR participants on our open-source dataset.

\section{Localized critical peak rebates in Québec}

Québec's CPR and LCPR programs have been carried out under the banner of Hilo, a division of Hydro-Québec in charge of DR aggregation. Hilo, calls a DR \textit{challenge} a day ahead of the event, and users choose their degree of participation for the next day. Hilo subscribers are equipped with smart thermostats and their respective heating setpoints are controlled by Hilo, according to an agreed-upon strategy, throughout the event. With the Hilo mobile application and different connected objects, customers can program the response of their house to different \textit{scenes}. For example, they can choose the heating setpoint of each thermostat when there is a DR event, or when nobody is home. When notified, smart thermostats to augment user comfort during events which typically last 4 hours: either between 6 AM and 10 AM or 5 PM and 9 PM. A maximum of 30 CPR events can be called per year. LCPR is an additional program currently under testing. Testers can be asked for up to 10 extra LCPR events. Rewards are proportional to the total amount of energy shaved during the event, i.e., heating-related curtailment and others, with respect to their estimated baselines~\cite{hilo_web}. 

\section{Open dataset}
\paragraph{Description} The dataset we share contains the aggregated hourly consumption of 197 anonymous LCPR testers located in three substations. Additional hourly weather data and LCPR information are also present. Table \ref{tab:features} details the features and the label. Note that we also provide cyclical encoding of temporal features, e.g., month, day of the week, and hour. We remark that outliers and anomalies are present in the dataset because of metering and telemetry issues or even blackouts, e.g., an aberrant (and impossible) 32.2 MWh energy consumption is registered at some point.
\begin{table}[tb]
\renewcommand{\arraystretch}{1.0}
    \centering
    \caption{Description of features and label of the dataset}
    \begin{tblr}{colspec={m{5cm}|m{5cm}|c},hline{1,25} = {1.5pt}, hline{2,3,4,5,6,7,8,9,10,11,12,13,14,15,16,17,18,19,20,21,22,23,24}={1pt}} 
    \textbf{Name} & \textbf{Description} & \textbf{Possible values}  \\
    \texttt{substation} & Substation identifier & \{\textquotesingle A\textquotesingle, \textquotesingle B\textquotesingle, \textquotesingle C\textquotesingle\}\\
    \texttt{timestamp\_local} & Timestamp in local time (UTC-5) and ISO 8601 format [AAAA-MM-DD hh:mm:ss]& $-$\\
    \texttt{connected\_clients} & Number of clients connected to the substation during the considered hour & $\{9,10,\ldots,104\}$ \\
    \texttt{connected\_smart\_tstats} & Number of smart thermostats connected to the substation during the considered hour & $\{59,60,\ldots,1278\}$\\
    \texttt{average\_inside\_temperature} & Hourly average indoor  temperature measured by smart thermostats in substation [$^\circ$C]& $[16.21,27.08]$ \\
    \texttt{average\_temperature\_setpoint} & Hourly average setpoint of smart thermostats in substation [$^\circ$C] & $[9.31, 21.03]$\\
    \texttt{average\_outside\_temperature} & Hourly average outside temperature at substation [$^\circ$C] & $[-32.0, 35.2]$ \\
    \texttt{average\_solar\_radiance} & Hourly average solar radiance at substation [W/m$^2$] & $[0,961]$ \\
    \texttt{average\_relative\_humidity} & Hourly average relative humidity at substation [\%] & [0,100]\\
    \texttt{average\_snow\_precipitation} & Hourly average amount of snow precipitation at substation [mm] & [0.0,306.0]\\
    \texttt{average\_wind\_speed} & Hourly average wind speed at substation [m/s] & [0, 15.68 ]\\
    \texttt{date} & Date [AAAA-MM-DD]& [2022-01-01, 2024-06-30]\\
    \texttt{month} & Month &  $\{1, 2, \ldots, 12\}$ \\
    \texttt{day} & Day of the month& $\{1, 2, \ldots, 31\}$\\
    \texttt{day\_of\_week} & Day of the week with Sunday and Saturday being 1 and 7, respectively&  $\{1, 2, \ldots, 7\}$\\
    \texttt{hour} & Hour of the day&  $\{0, 1, \ldots, 23\}$\\
     \texttt{challenge\_type} & Type of challenge during the given hour & \{\textquotesingle None\textquotesingle, \textquotesingle CPR\textquotesingle, \textquotesingle LCPR\textquotesingle\}\\
      \texttt{challenge\_flag} & Flag indicating hours in challenge&  $\{0, 1\}$ \\
      \texttt{pre\_post\_challenge\_flag} & Flag indicating hours in pre-challenge or post-challenge&  $\{0, 1\}$ \\
    \texttt{is\_weekend} & Flag indicating weekends & $\{0, 1\}$\\
     \texttt{is\_holiday} & Flag indicating Québec holidays  &  $\{0, 1\}$\\
      \texttt{weekend\_holiday} & Flag indicating whether a weekend or a holiday &  $\{0, 1\}$\\
      \texttt{total\_energy\_consumed } &Hourly energy consumption of the substation [kWh]  &  $[7.45, 32240.17]$\\
\end{tblr}
    \label{tab:features}
\end{table}
We refer readers to Appendix \ref{app:data} for additional analyses and visualizations of the dataset's features and labels.

\paragraph{Open data initiatives at Hydro-Québec}
The yearly energy demand in Québec should double by 2050, requiring an additional production of 60 TWh by 2035 and the addition of 8 to 9 GW of peak power to Québec's 38 GW capacity~\cite{Hydro-Québec_2023a}. Québec's energy landscape must evolve rapidly to meet the increasing demand and DR is a powerful tool to mitigate infrastructure growth.

Hydro-Québec recognizes the importance of sharing data to foster research, innovation, and informed decision-making. Through its plateform\footnote{\url{https://www.hydroquebec.com/documents-data/open-data/}}, it provides historical and real-time data on energy production and consumption in Québec as well as other datasets, e.g., geolocated hydrometeorological data from its remote weather stations. This new dataset is Hydro-Québec's first dedicated to DR research.

By supporting data democratization, Hydro-Québec encourages the emergence of a culture of transparency and openness in the energy sector which is needed to accelerate the energy transition. This approach aims to stimulate researchers, citizens, and businesses in their desire to participate in the design of innovative applications for a more sustainable, interoperable, reliable, and safe power~grid.

\section{Sliced-Wasserstein Filter}

The Wasserstein distance is a metric that provides a sense of distance between two distributions. Also called earth mover's distance, it can be conceived as the minimal effort it would take to displace a pile of a weighted resource to shape a specific second pile~\cite{panaretos2019statistical}. The order-$t$ Wasserstein distance with respect to some norm $\|\cdot\|$ between distributions $\mathbb{U}$ and $\mathbb{V}$ is defined as: $$W_{\|\cdot\|,t}(\mathbb{U}, \mathbb{V}) = \left( \min_{\pi \in \mathcal{J}(\mathbb{U}, \mathbb{V})} \int_{\mathcal{Z}\times\mathcal{Z}} \|\mathbf{z}_1 - \mathbf{z}_2\|^t \text{d}\pi(\mathbf{z}_1,\mathbf{z}_2)\right)^{1/t},$$
where $\mathcal{J}(\mathbb{U}, \mathbb{V})$ represents all the possible joint distributions $\pi$ between $\mathbb{U}$ and $\mathbb{V}$, $\mathbf{z}_1$ and $\mathbf{z}_2$ are the marginals of $\mathbb{U}$ and $\mathbb{V}$, respectively, and $\mathcal{Z}$ is the set of all possible values of $\mathbf{z}_1$ and $\mathbf{z}_2$~\cite{Chen2020_DRO}. In general, computing the Wasserstein distance is of high computational complexity \cite{kolouri2019generalized} except for some special cases. For example, the Wasserstein distance for one-dimensional distributions has a closed-form solution that can be efficiently approximated. The sliced-Wasserstein (SW) distance is a metric that makes use of this property by calculating infinitely many linear projections of the high-dimensional distribution onto one-dimensional distributions and then computing the average of the Wasserstein distance between these one-dimensional representations \cite{bonneel2015sliced, kolouri2019generalized}. Interestingly, it possesses similar theoretical properties to the original Wasserstein distance while being computationally tractable \cite{bonnotte2013unidimensional}. The order-$t$ SW distance's approximation under a Monte Carlo scheme is defined as: 
\begin{equation*}S W_{\|\cdot\|,t}\left(\mathbb{U}, \mathbb{V} \right) \approx\left(\frac{1}{L} \sum_{l=1}^L W_{\|\cdot\|,t}\left(\mathcal{R} \mathbb{U}\left(\cdot, \theta_l\right), \mathcal{R} \mathbb{V}\left(\cdot, \theta_l\right)\right)^t\right)^{1 / t},\end{equation*} 
where $\mathcal{R} \mathbb{D}(\cdot, \boldsymbol{\theta}_l)$ is a single projection of the Radon transform of distribution function $\mathbb{D}$ over a fixed sample $\boldsymbol{\theta}_l \in \mathbb{S}^{d-1} = \{\boldsymbol{\theta} \in \mathbb{R}^d | \theta_1^2 + \theta_2^2 + \cdots + \theta_d^2 = 1\}$ $\forall l \in \llbracket L \rrbracket$. Interested readers are referred to references \cite{bonneel2015sliced, kolouri2019generalized} for in-depth mathematical explanations.

We now utilize this approximation of the sliced-Wasserstein distance to formulate a simple outlier filter. Consider an empirical distribution $\hat{\mathbb{P}}_N = \frac{1}{N} \sum_{i=1}^N \delta_{\mathbf{z}_i}(\mathbf{z})$, where $\delta_{\mathbf{z}_i}(\cdot)$ is a Dirac delta function assigning a probability mass of one at each known sample $\mathbf{z}_i \in \mathbb{R}^d$. Consider the notation $\hat{\mathbb{P}}_{N-1}^{-\mathbf{z}_i}$ to denote a variation of $\hat{\mathbb{P}}_N$ in which we remove sample $\mathbf{z}_i$. As the weight of each sample is identical and because the SW distance compares distributions of equal weights, we propose an outlier filter by using a simple voting system in which we compare the SW distance between the empirical distribution minus the outlier candidate and the empirical distribution minus a random sample. Let $\mathcal{O} \subseteq \mathcal{D}$ and $\mathcal{O}^c \subseteq \mathcal{D}$ denote the sets of outlier and inlier samples, respectively, under the perspective of the filter and let $\mathcal{D} = \mathcal{O} \cup \mathcal{O}^c \subset \mathbb{R}^d$ be the available dataset. A vote is positive if the distance between the two empirical distributions exceeds the threshold $\epsilon > 0$. A sample is labelled an outlier if the proportion of positive votes is greater than the threshold $p$.
\begin{equation}\label{eq:swod}\tag{\texttt{SWAD}}
    \mathcal{O} = \left\{ \mathbf{z}_i \in \mathcal{D} \ \middle\vert \ p \leq \frac{1}{n} \sum_{\mathbf{z}_j \sim \hat{\mathbb{P}}_{N-1}^{-\mathbf{z}_i}: j\in \llbracket n \rrbracket} \mathbbm{1}\left( SW_{\|\cdot\|,t} (\hat{\mathbb{P}}_{N-1}^{-\mathbf{z}_i},\hat{\mathbb{P}}_{N-1}^{-\mathbf{z}_j} ) \geq \epsilon \right),  \forall i \in \llbracket N\rrbracket \right\},
\end{equation}
where $\mathbbm{1}(\cdot)$ denotes an indicator function, $n$ is the number of points used for the vote, $p \in [0,1]$ is the voting threshold required to label a sample as an outlier, and $\llbracket N \rrbracket \equiv \{1, 2, \ldots, N\}$. 

This filter is interesting for several reasons. It is unsupervised, purely analytical, and uses a well-known explainable mathematical distance to filter data points that seem out-of-sample under the SW metric. The intuition is that we remove samples that are costly in the transportation plan when compared to other random samples of the same distribution. We can use the voting percentage to measure the algorithm's confidence in its labelling. It is also parallelizable, as seen in our implementation provided on our GitHub page\footnote{\url{https://github.com/jupall/swfilter}}. Numerical results and figures are presented in Appendix \ref{app:SW}.

This method does not scale with large datasets as is: the SW distance computational burden increases as bigger distributions are compared. We propose a smart splitting method to accelerate the procedure, but we remark that the transportation plan between a distribution minus a sample and the same distribution minus another sample can be roughly approximated by the Euclidian distance between the two removed samples. As such, we introduce a fast Euclidian approximation for $W_{\|\cdot\|_2,1}$:
\begin{equation}\label{eq:fead}\tag{\texttt{FEAD}}
    \mathcal{O} \approx \left\{ \mathbf{z}_i \in {D} \ \middle\vert \ p \leq \frac{1}{n} \sum_{\mathbf{z}_j \sim \hat{\mathbb{P}}_{N-1}^{-\mathbf{z}_i}}\mathbbm{1}\left( \|\mathbf{z}_i - \mathbf{z}_j\|_2 \geq \eta \right), \forall i \in \llbracket N\rrbracket \right\},
\end{equation} 
where $\eta$ is the threshold of the Euclidian distance. This method is not as accurate as the first proposal to filter out-of-sample data points. But, because (\ref{eq:swod}) and (\ref{eq:fead}) use the same principle, they share similar classification patterns when $\epsilon$ and $\eta$ are tuned accordingly, as can be seen in Figure \ref{fig:2d_toy}.

\section{Benchmark}
We now propose a first benchmark on our LCPR dataset. Our goal is to predict the aggregated hourly consumption at each substation in winter when peak demand is critical. To follow the literature~\cite{weng2015GP}, and propose a simple yet meaningful benchmark, we implement a Gaussian process~\cite{williams1995gaussian} in a rolling horizon fashion. Samples dating from before 2023-12-15 are used for hyperparameter tuning while those between 2023-12-15 and 2024-04-15, during the most recent winter DR season, are used for testing. We train one model per week and the training window size is a hyperparameter to be tuned. The SW filter is used to preprocess data in the tuning phase and increase generalization in testing. Our method is interesting because it can filter clear out-of-sample points while avoiding sparse LCPR events that other methods could consider as local outliers. We refer interested readers to our GitHub page for more details. Testing mean average errors (MAE) and root mean squared errors (RMSE) are presented in Table \ref{tab:bench} for each substation. See Appendix \ref{app:bench} for additional figures.

\section{Closing remark}
In this work, we present a new unsupervised outlier filter by leveraging the sliced-Wasserstein metric. This filter is interesting for MLOps integration on applications where global outliers may be adversarial to the prediction quality of trained models, e.g., smart grid data. We also hope to stimulate research on trustworthy ML models in critical sectors by releasing the first open dataset showcasing localized critical peak rebate demand response schemes in a northern climate. This dataset has a strong potential for benchmarking of such models as it opens a window to a real-world critical application where accuracy and robustness are equally important. To get the ball rolling, we provide a first benchmark by tuning simultaneously our SW filter and a Gaussian process.

\bibliographystyle{IEEEtran}
\bibliography{ref.bib}

\appendix
\section*{Appendices}
\addcontentsline{toc}{section}{Appendices}
\renewcommand{\thesubsection}{\Alph{subsection}}
\subsection{Detailed analysis of the dataset}\label{app:data}
\begin{figure}[h]
    \centering
    \includegraphics[width=0.95\textwidth]{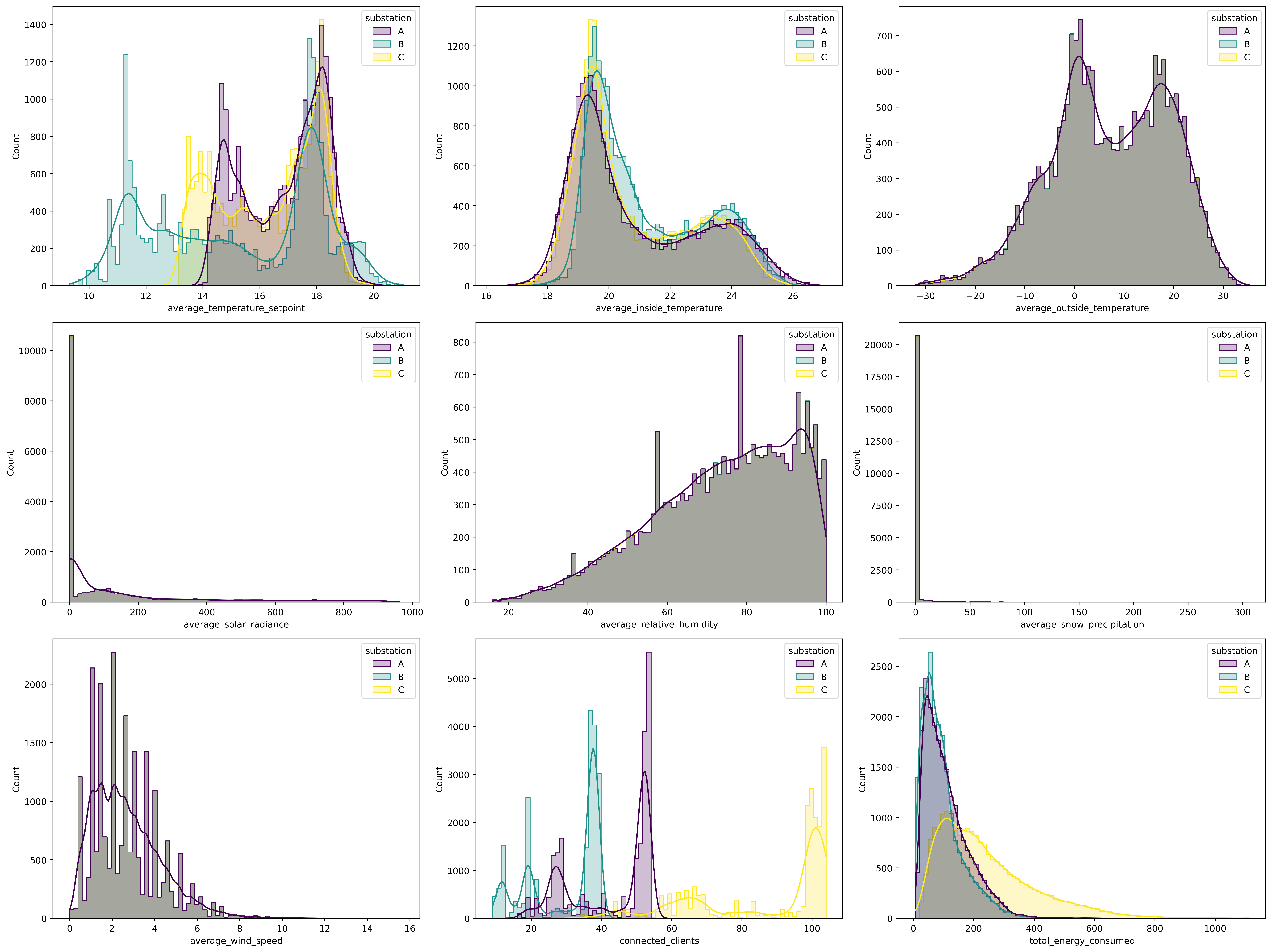}
    \caption{Distribution of key features for each substation}
    \label{fig:dist}
\end{figure}

In this section, we present some additional insights on the LCPR dataset. Figure \ref{fig:dist} presents the distribution count of some key features for each substation. We observe that meteorological features are identical for each substation as they are geographically adjacent and located in dense neighborhoods. 

\begin{figure}[h]
    \centering
    \begin{subfigure}{0.48\textwidth}
        \includegraphics[width=\textwidth]{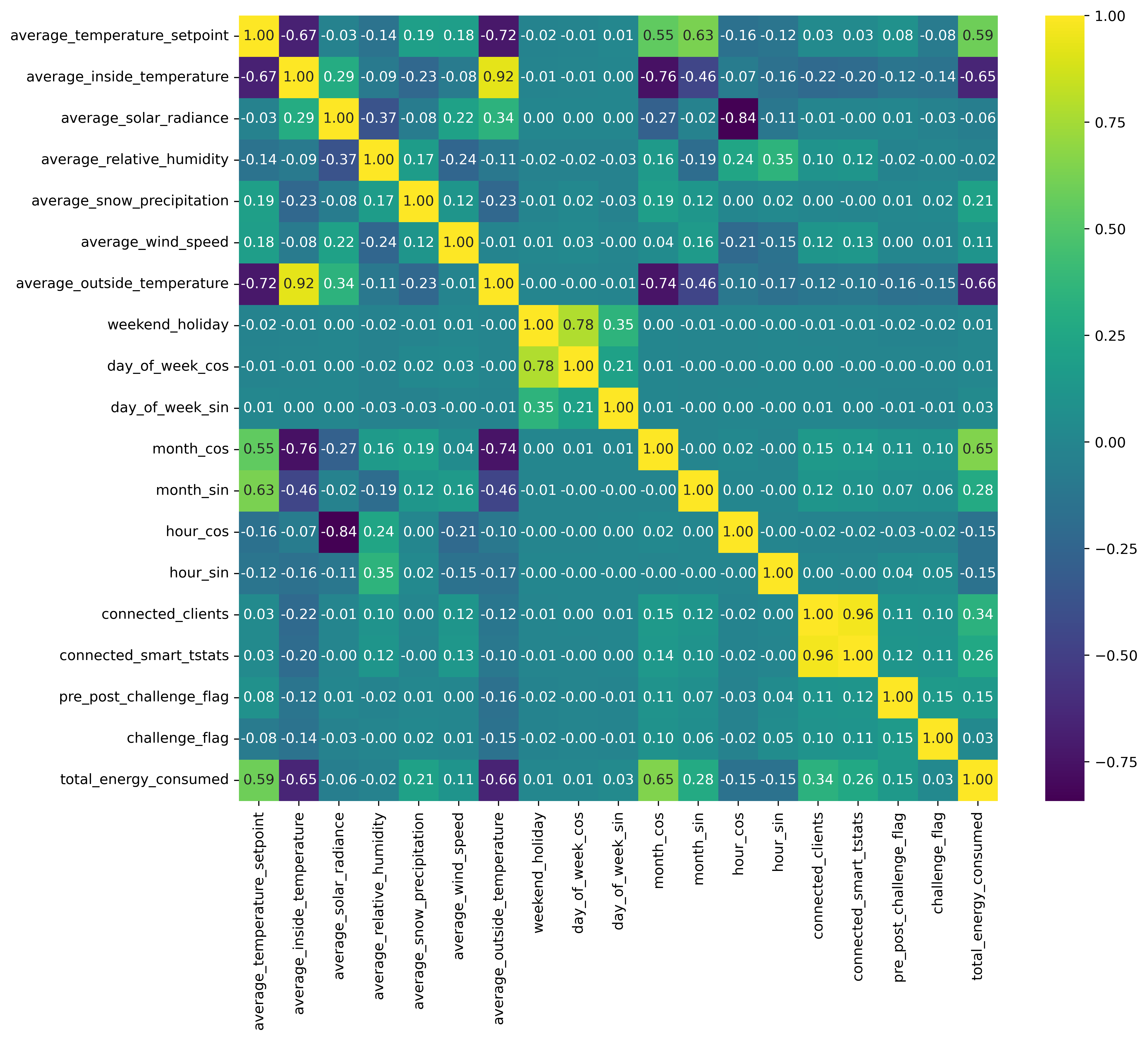}
        \caption{Substation A}
        %\label{fig:sub1}
    \end{subfigure}
    \hfill
    \begin{subfigure}{0.48\textwidth}
        \includegraphics[width=\textwidth]{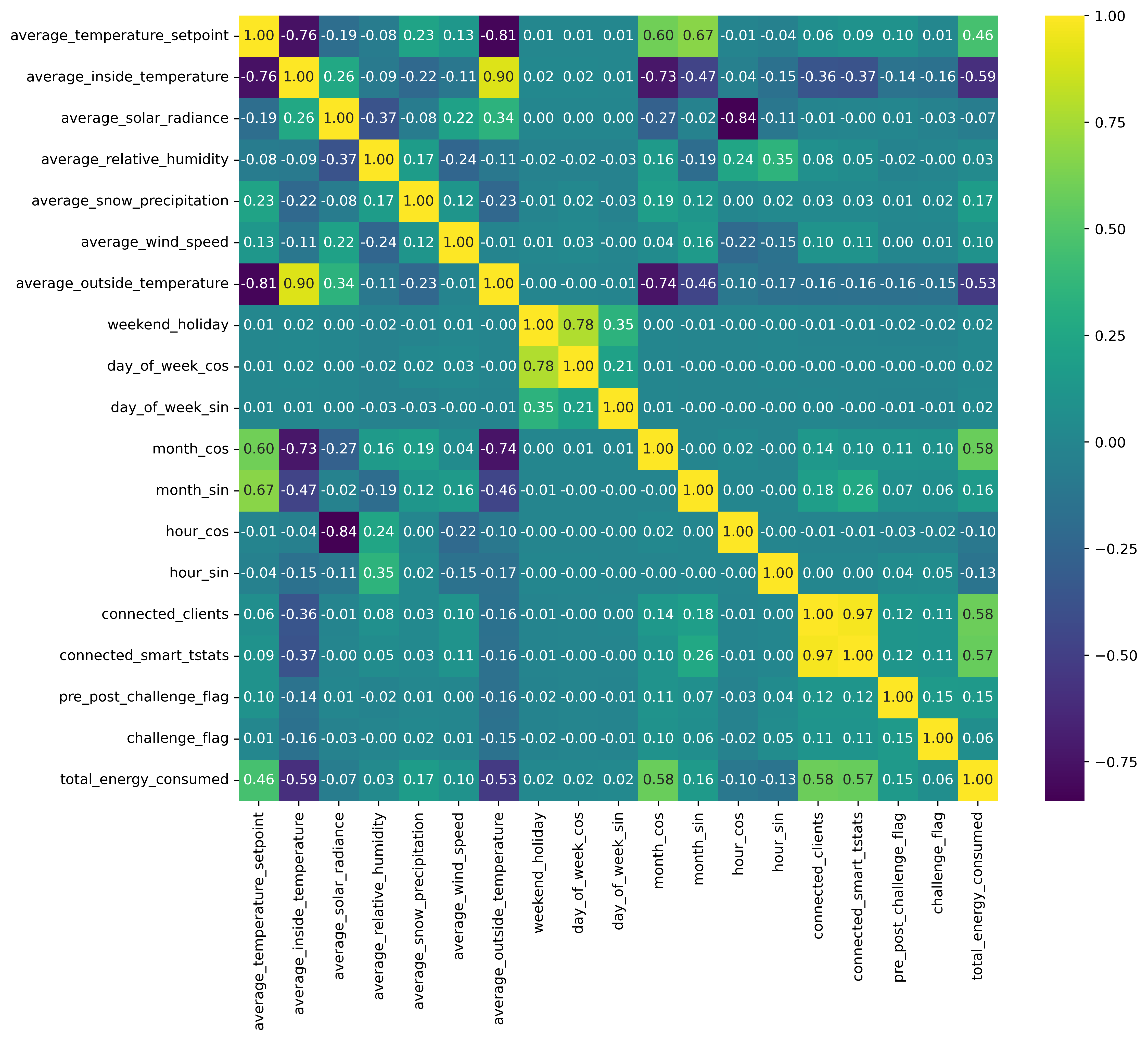}
        \caption{Substation B}
        %\label{fig:sub2}
    \end{subfigure}
    \hfill
        \begin{subfigure}{0.48\textwidth}
        \includegraphics[width=\textwidth]{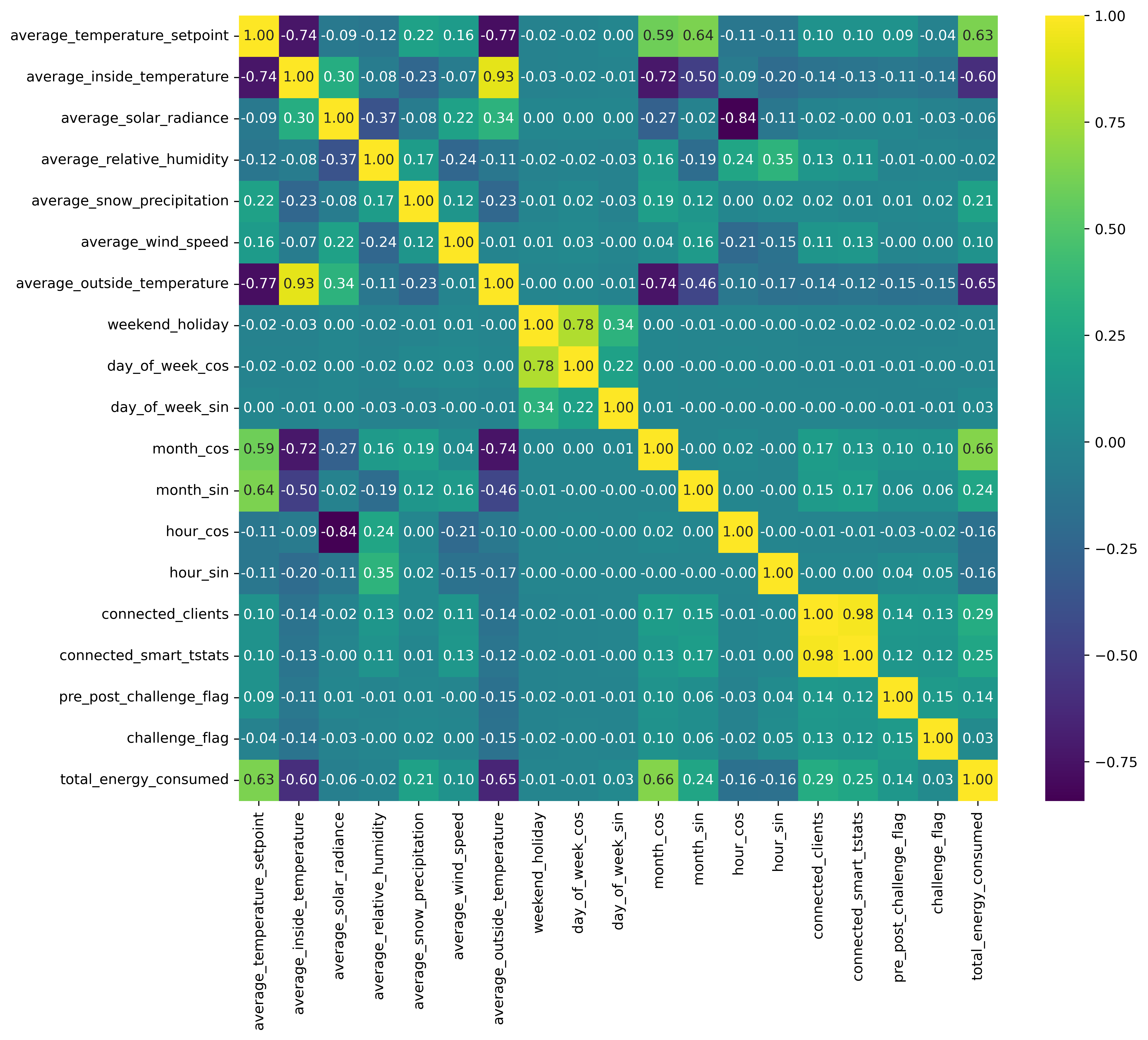}
        \caption{Substation C}
        %\label{fig:sub1}
    \end{subfigure}
    \caption{Correlation heatmap of key features and label for each substation}
    \label{fig:corr}
\end{figure}

\begin{figure}[tb]
    \centering
    \includegraphics[width=0.75\textwidth]{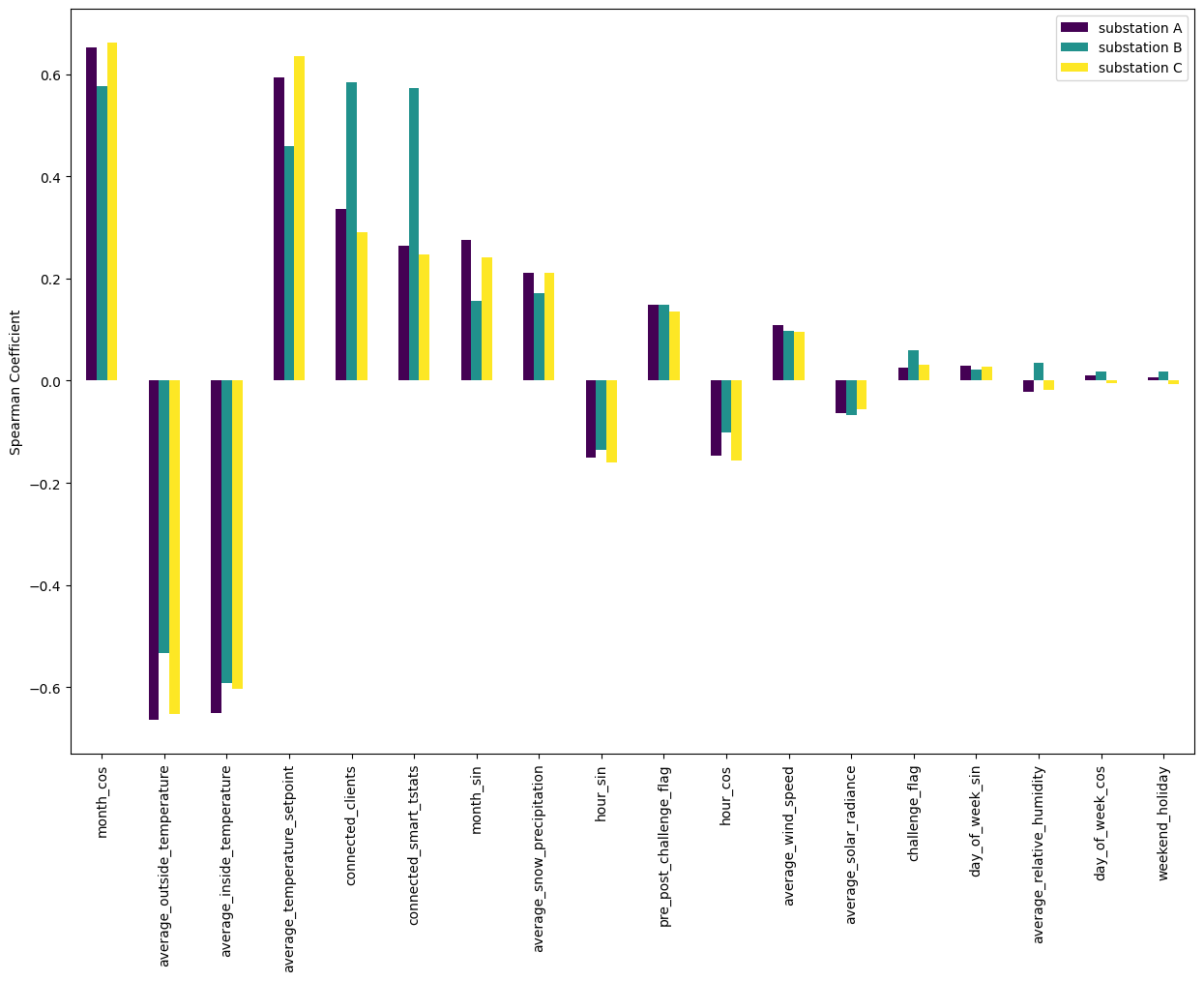}
    \caption{Spearman coefficients of key features for each substation}
    \label{fig:spear}
\end{figure}

Figure \ref{fig:corr} shows a correlation heatmap of important features and label for each substation. We observe that each substation follows the same general tendencies. We remark significant correlations between the energy consumed, the month of the year, the outside temperatures, and the temperature setpoints. This is also highlighted in Figure \ref{fig:spear} which presents the Spearman coefficients ranking~\cite{spearman04} between each feature and the label. A positive sign indicates that both the label and the feature grow or decrease in the same direction while a negative sign indicates an opposite direction. The coefficients are ranked in decreasing importance from left to right. 

\begin{figure}[tb]
    \centering
    \begin{subfigure}{0.48\textwidth}
        \includegraphics[width=\textwidth]{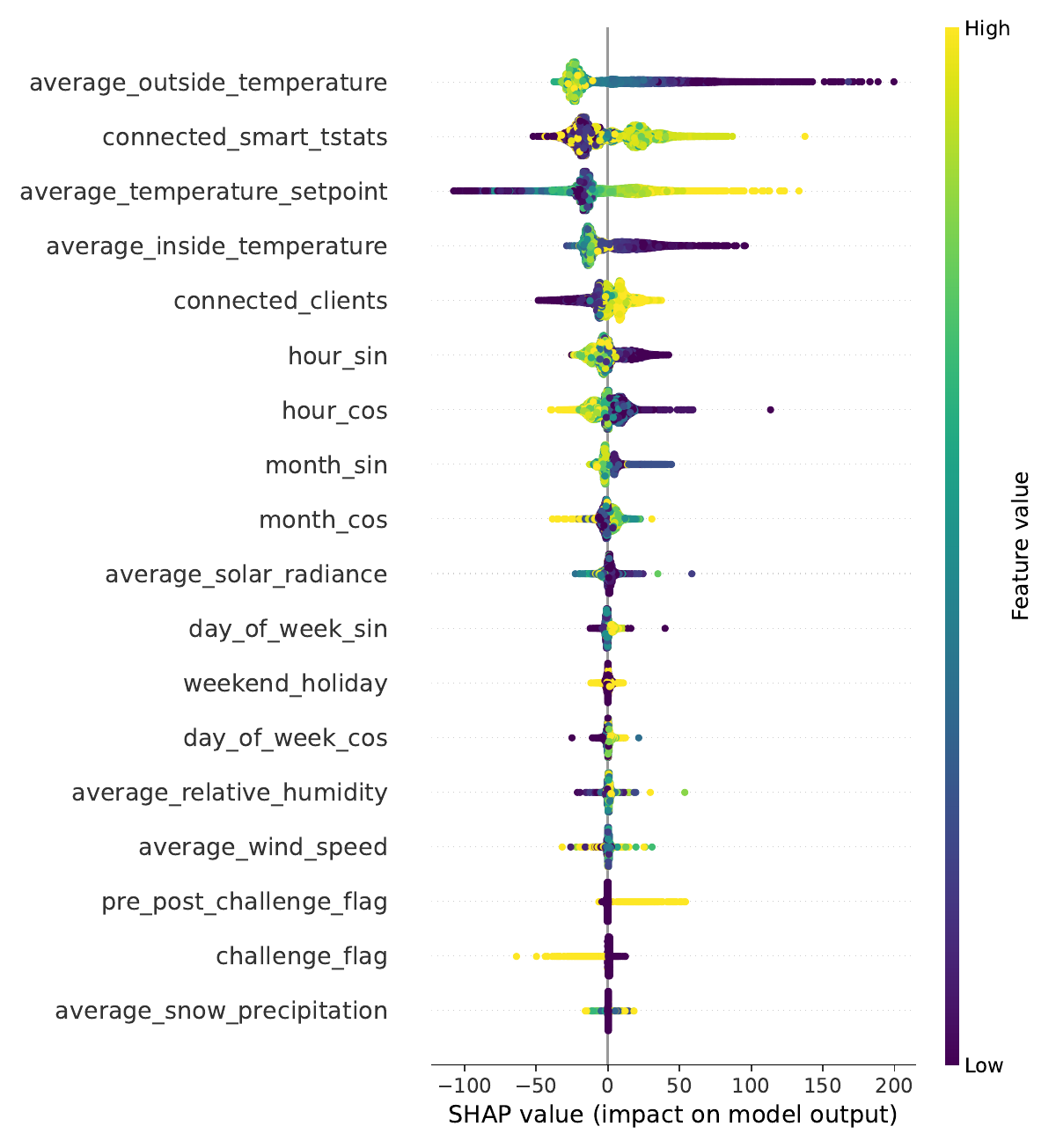}
        \caption{Substation A}
        %\label{fig:sub1}
    \end{subfigure}
    \hfill
    \begin{subfigure}{0.48\textwidth}
        \includegraphics[width=\textwidth]{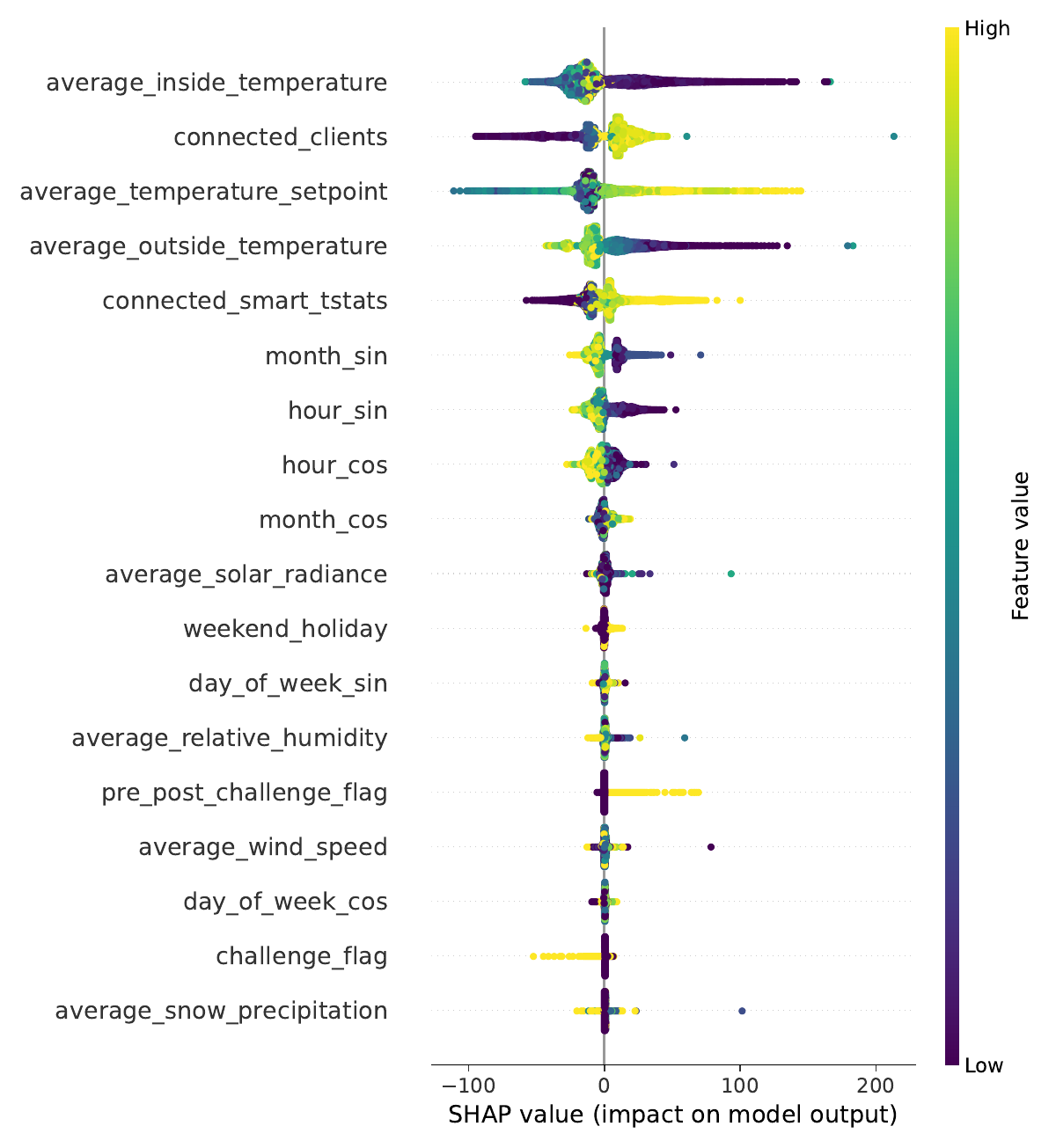}
        \caption{Substation B}
        %\label{fig:sub2}
    \end{subfigure}
    \hfill
        \begin{subfigure}{0.48\textwidth}
        \includegraphics[width=\textwidth]{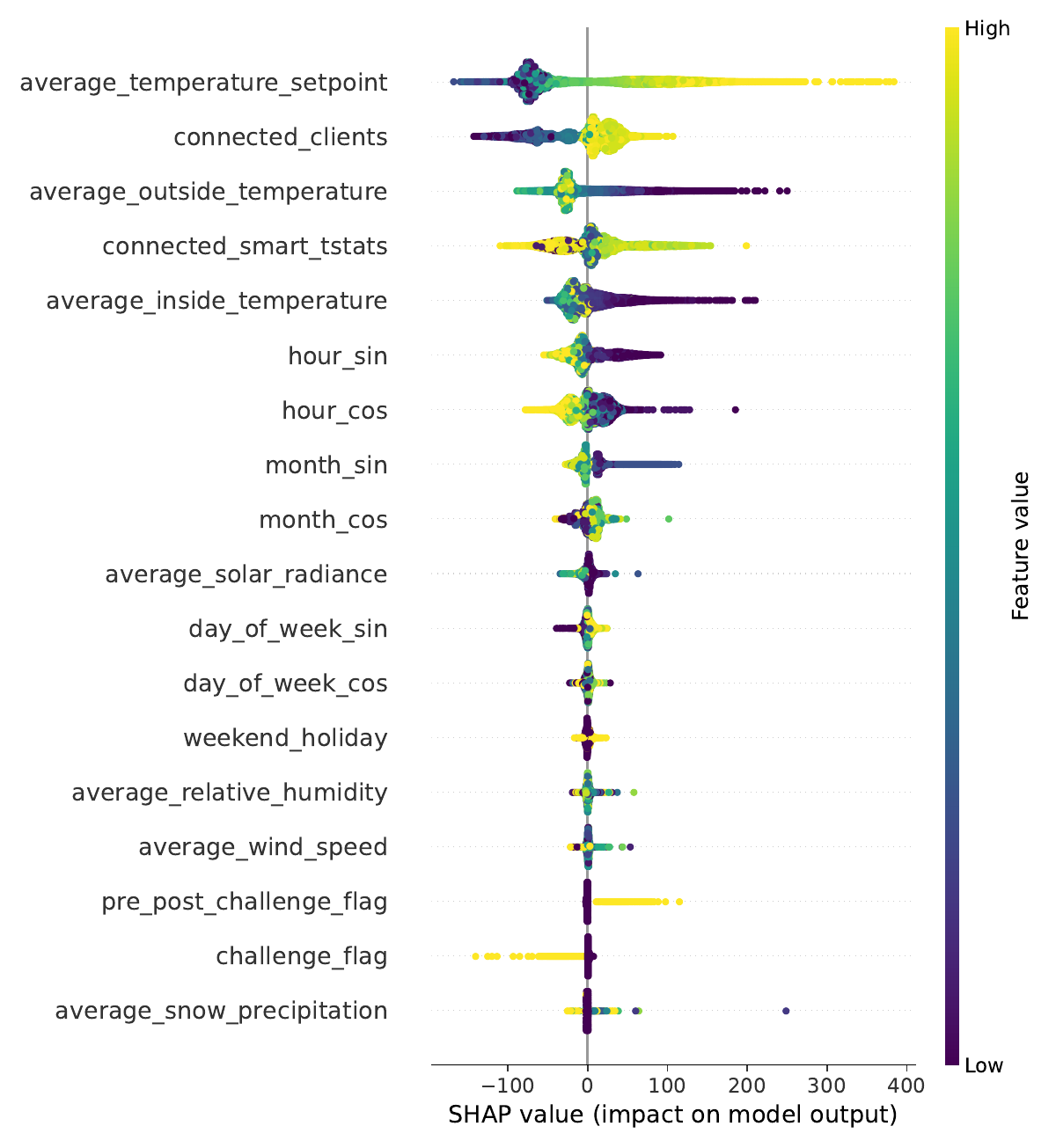}
        \caption{Substation C}
        %\label{fig:sub1}
    \end{subfigure}
    \caption{Shapley analysis of key features for each substation on trained \texttt{XGBoost}}
    \label{fig:shap}
\end{figure}

To have a more nuanced analysis of the contribution of each feature to the output, we also provide in Figure \ref{fig:shap} an analysis of Shapley values of a trained extreme gradient boosting model (\texttt{XGBoost})~\cite{chen2016xgboost} for each substation. These analyses were realized with the Python package \texttt{SHAP}~\cite{lundberg2017SHAP}. As we see, some lower-ranked features, viz., the challenge flags, sometimes have a strong impact on the model's output even though their general impact is null.

\subsection{Numerical study of the Sliced-Wasserstein Filter}\label{app:SW}
\begin{figure}[tb]
    \centering
    \includegraphics[width=0.5\textwidth]{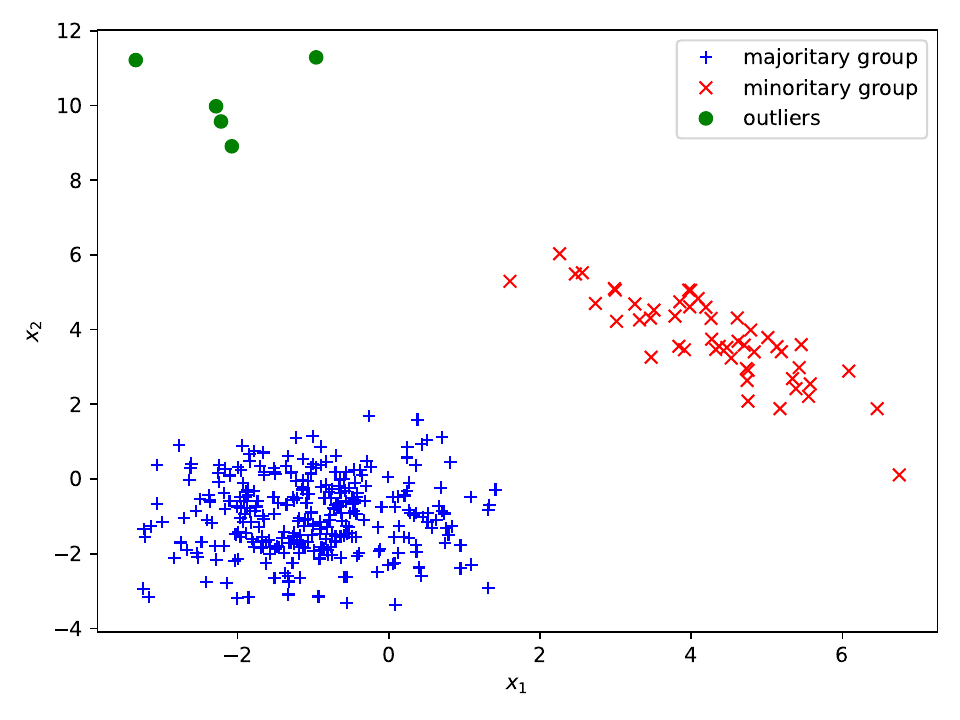}
    \caption{Illustration of different groups}
    \label{fig:2d_tut_dist}
\end{figure}

\begin{figure}[tb]
    \centering
    \begin{subfigure}{0.32\textwidth}
        \includegraphics[width=\textwidth]{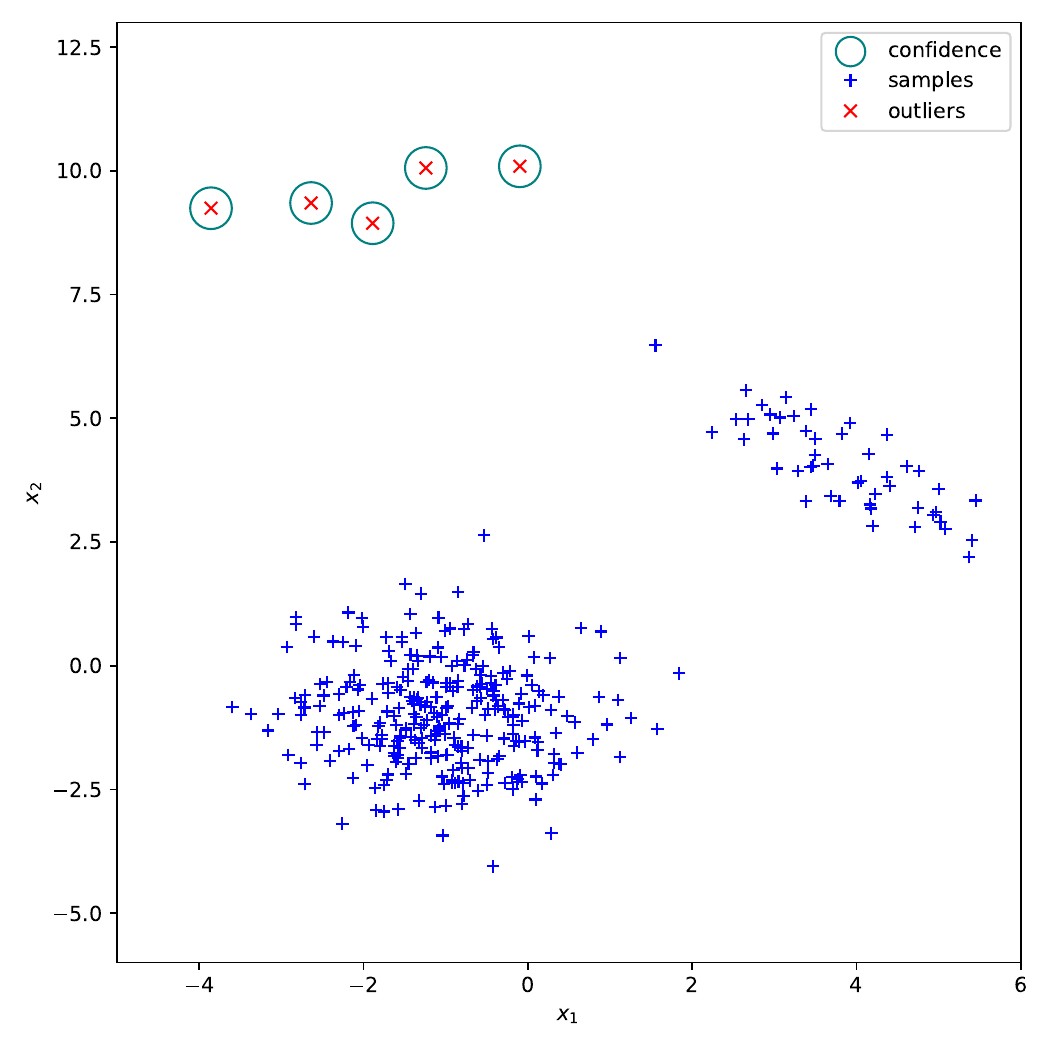}
        \caption{$\epsilon=0.1$}
        %\label{fig:sub1}
    \end{subfigure}
    \hfill
    \begin{subfigure}{0.32\textwidth}
        \includegraphics[width=\textwidth]{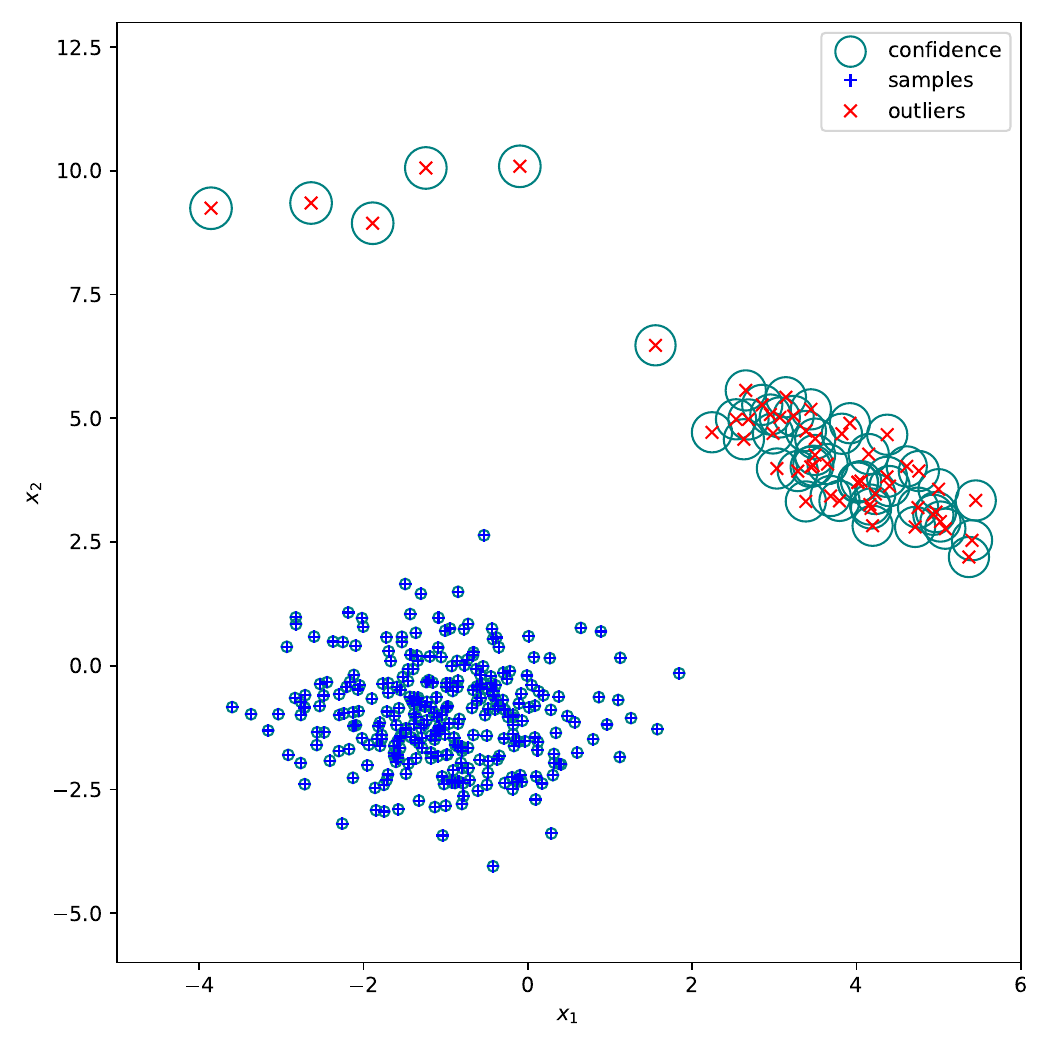}
        \caption{$\epsilon=0.05$}
        %\label{fig:sub2}
    \end{subfigure}
    \hfill
        \begin{subfigure}{0.32\textwidth}
        \includegraphics[width=\textwidth]{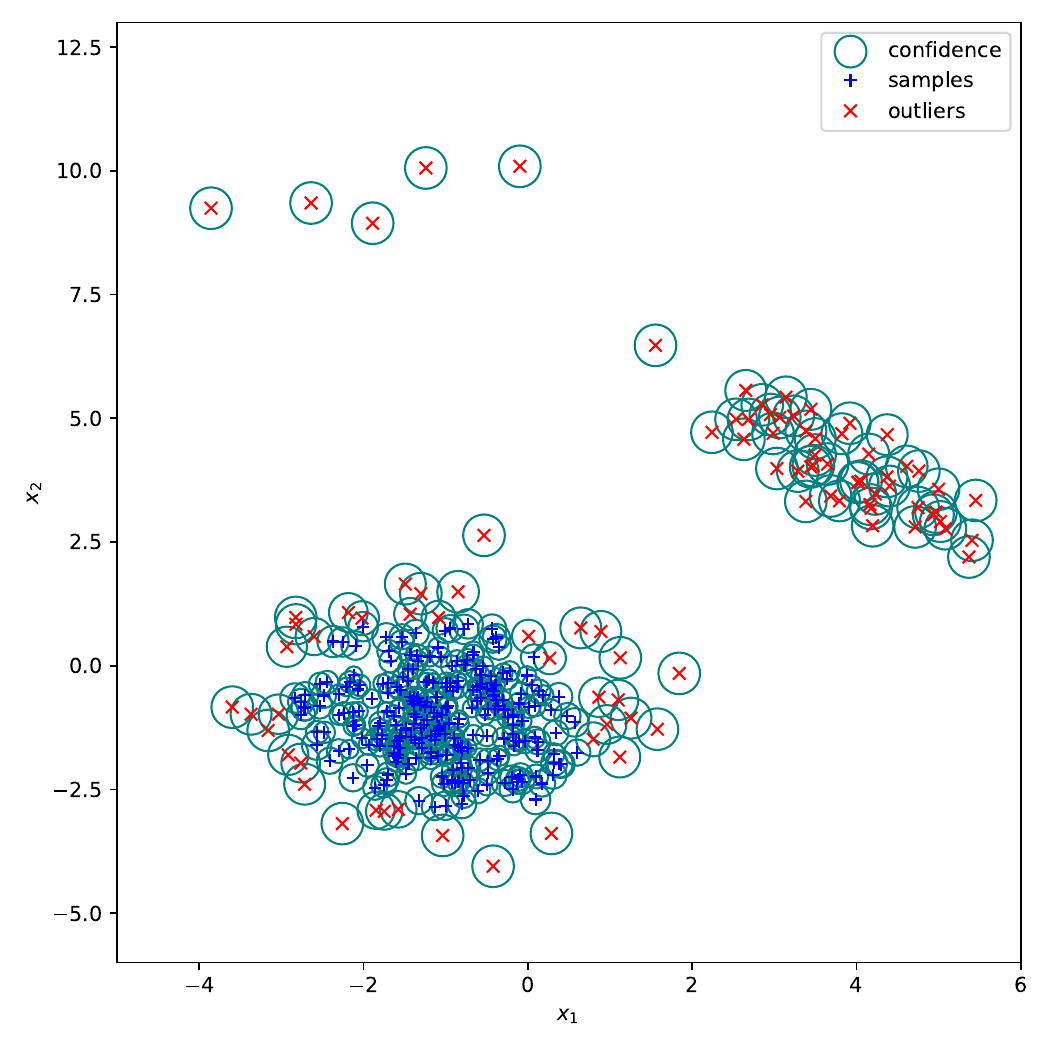}
        \caption{$\epsilon=0.01$}
        %\label{fig:sub1}
    \end{subfigure}
    \caption{Labelling of the SW filter for different values of $\epsilon$}
    \label{fig:2d_tut}
\end{figure}

We first begin this section by showing the AD mechanism of the SW filter on a simple two-dimensional example. We generate three Gaussian distributions with different population sizes. As shown in Figure \ref{fig:2d_tut_dist}, the first distribution is the majority group, the second is the minority group, and the third represents clear statistical outliers.
We now merge the three distributions into a single one to test our SW filter. We vary the radius of the Wasserstein ball to see how the filter behaves. The results are presented in Figure \ref{fig:2d_tut}.

%\begin{figure}[tb]
%    \centering
%    \includegraphics[width=\textwidth]{2D_tutorial.pdf}
%    \caption{Labeling of the SW filter for different values of $\epsilon$}
%    \label{fig:2d_tut}
%\end{figure}
As we see, we can generate three filtering scenarios by modifying the value of $\epsilon$. With $\epsilon = 0.1$, we filter only the statistical outliers. With $\epsilon = 0.05$, we only keep the majority group. And, with $\epsilon = 0.01$, we only keep the samples closest to the barycenter of the majority group. This is very interesting in a safe ML pipeline as we can tune the \textit{conservatism} of the training dataset that is used at each run during hyperparameter optimization.

\begin{figure}[tb]
    \centering
    \includegraphics[width=\textwidth]{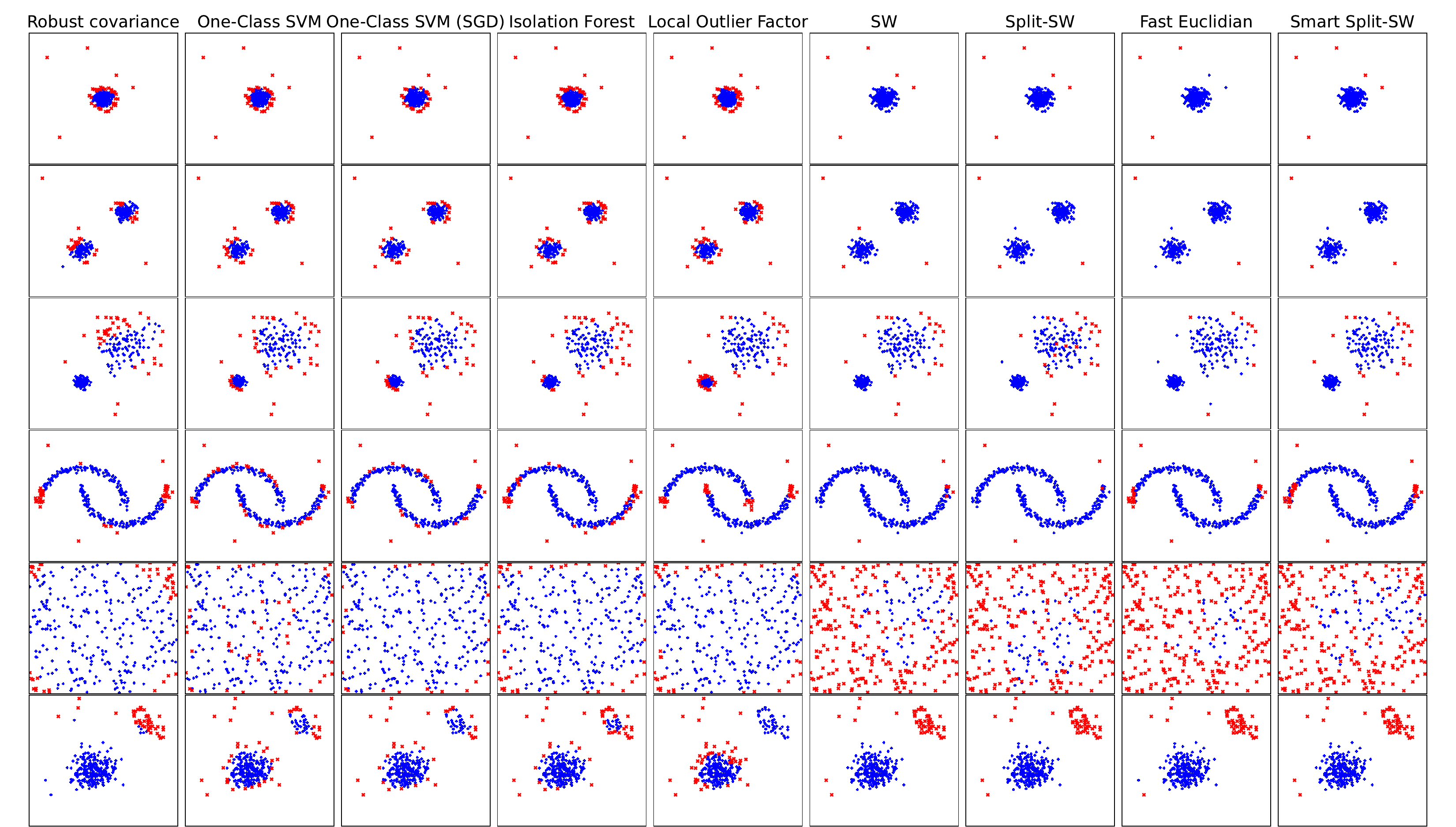}
    \caption{AD comparison on multiple synthetic datasets}
    \label{fig:2d_toy}
\end{figure}

In Figure \ref{fig:2d_toy}, we compare different AD algorithms on synthetic datasets provided in \texttt{scikit-learn}'s example collection~\cite{scikit-learn}. Hyperparameters are fixed for each algorithm to see how a single hyperparameter choice influences the labelling on each dataset. 
As can be seen, the SW filter and its fast Euclidian approximation are better at isolating outliers when there is a clear majority group but are not as precise in identifying local outliers based on local density, e.g., one-class SVM.

\begin{figure}[tb]
    \centering
    \begin{subfigure}{0.8\textwidth}
        \includegraphics[width=\textwidth]{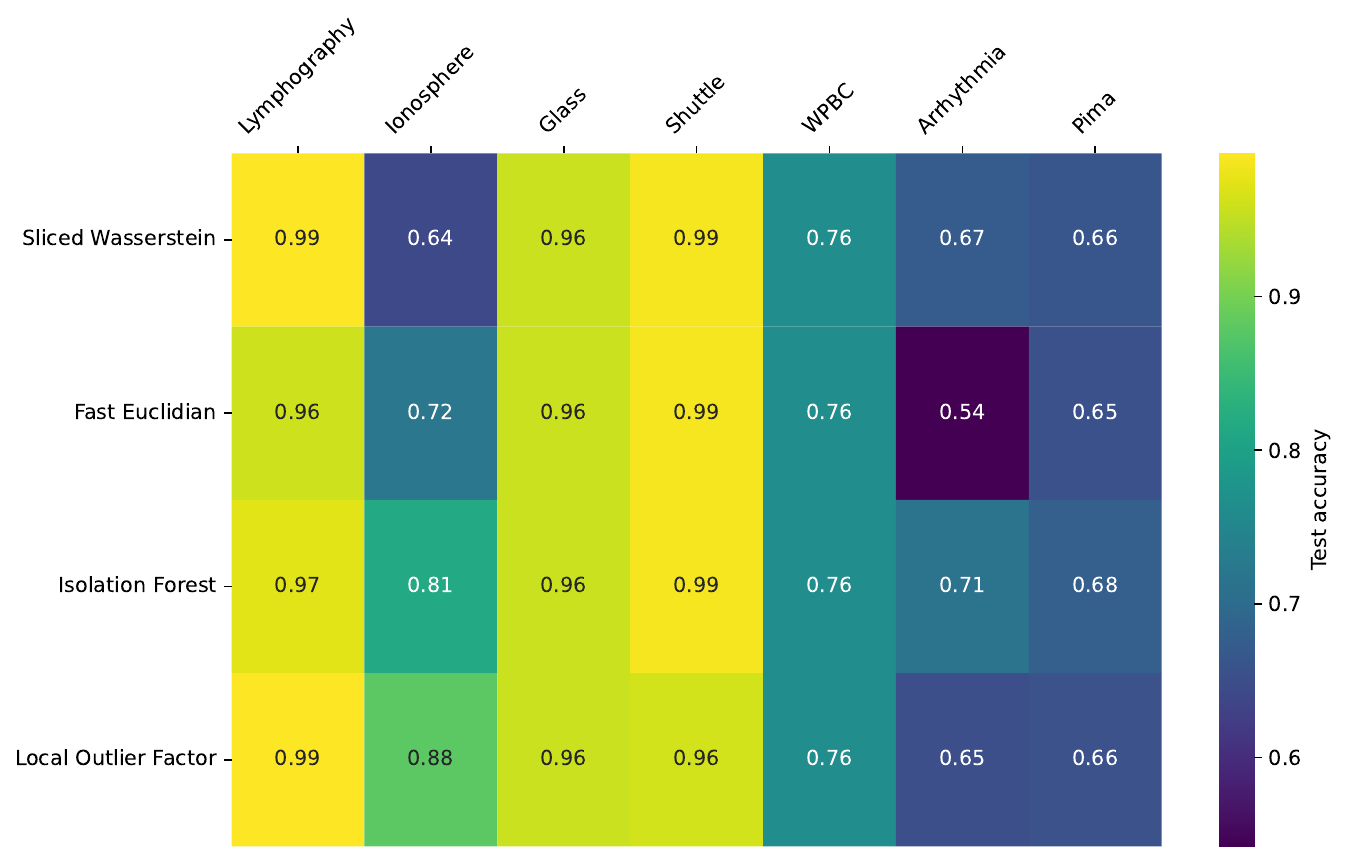}
        \caption{Accuracy $A$}
        %\label{fig:sub1}
    \end{subfigure}
    \hfill
    \begin{subfigure}{0.8\textwidth}
        \includegraphics[width=\textwidth]{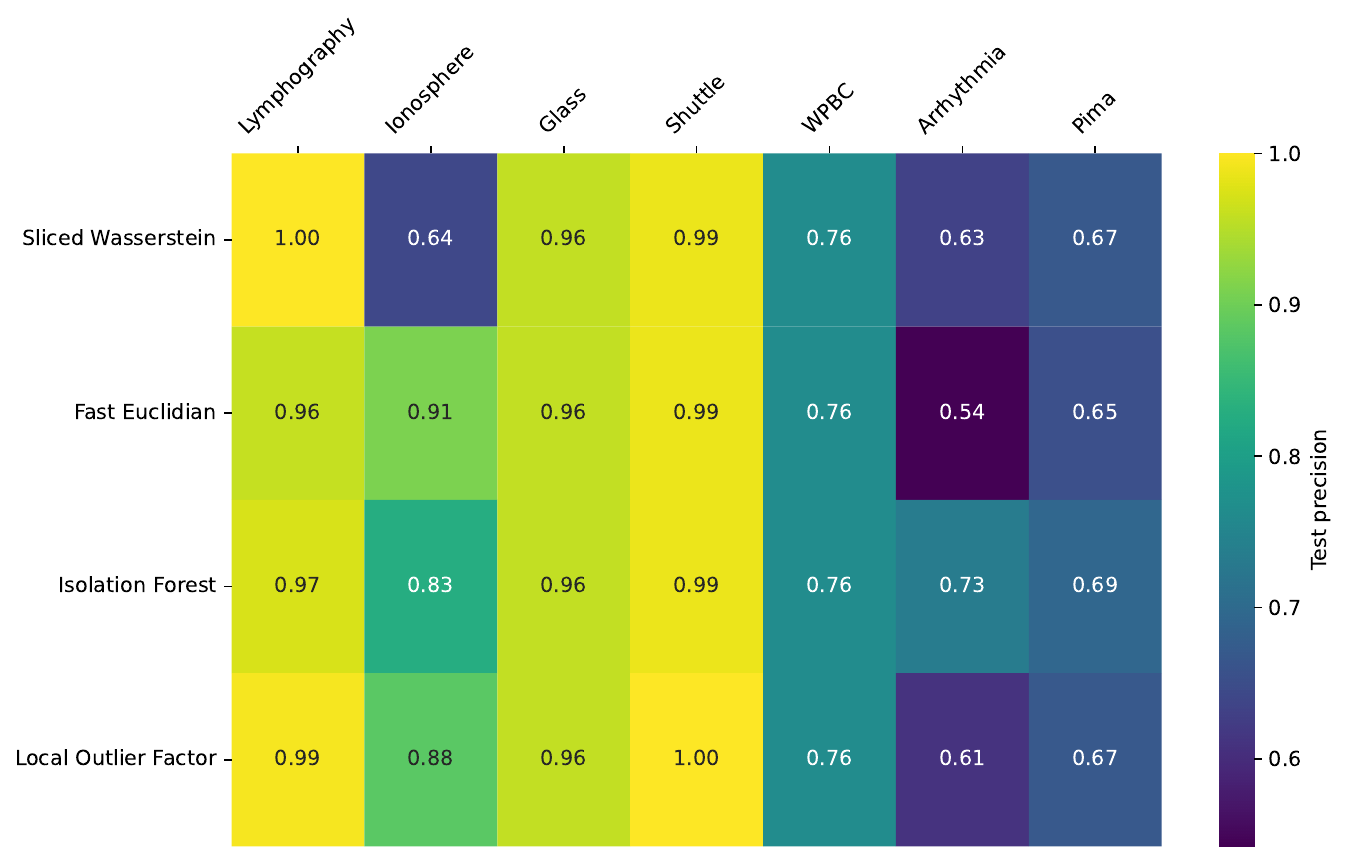}
        \caption{Precision $P$}
        %\label{fig:sub2}
    \end{subfigure}
    \caption{Results of the grid search for each AD model on each dataset}
    \label{fig:grid}
\end{figure}

Finally, we run a more thorough experiment with typically used real-world benchmark datasets. We run experiments on the \textit{Lymphography}, \textit{Ionosphere}, \textit{Glass}, \textit{Shuttle}, \textit{WPBC}, \textit{Arrhythmia}, and \textit{Pima}, datasets presented in \cite{campos2016evaluation} for AD benchmarking. We compare isolation forest and LOF to our method. We implement a grid search and select each model's run with the best accuracy $A$. We then extract the precision score $P$ from this run. The performance indicators are defined as follows:

$$P = \frac{T_\text{p}}{T_\text{p} + F_\text{p}}, \qquad A = \frac{T_\text{p} + T_\text{n}}{T_\text{p} + F_\text{p} + T_\text{n} + F_\text{n}}, $$
where $T_\text{p}$, $F_\text{p}$, $T_\text{n}$, and $F_\text{n}$ stand for true positives, false positives, true negatives, and false negatives, respectively. Results are presented in Figure \ref{fig:grid}. We observe a similar performance between each model, except on the \textit{Ionosphere} dataset where the SW filter lags behind the other models, and on \textit{Arrhythmia} where the fast Euclidian approximation is underperforming. The SW filter's strength is that it considers the global distributional properties of the population to guide its labelling. We remark that our method fails at detecting local outliers as it is purely designed to locate global outliers that are \textit{risky} to keep in the training set. This is a reasonable choice to make when designing data preprocessing methods for safer ML training but not necessarily to filter local outliers.

Every experiment is available on our GitHub page.

\subsection{Supplementary content to the benchmark}\label{app:bench}
This section presents visual test predictions of the best validation runs for each substation as well as the test error of the benchmark.

\begin{figure}[tb]
    \centering
    \begin{subfigure}{0.99\textwidth}
        \includegraphics[width=\textwidth]{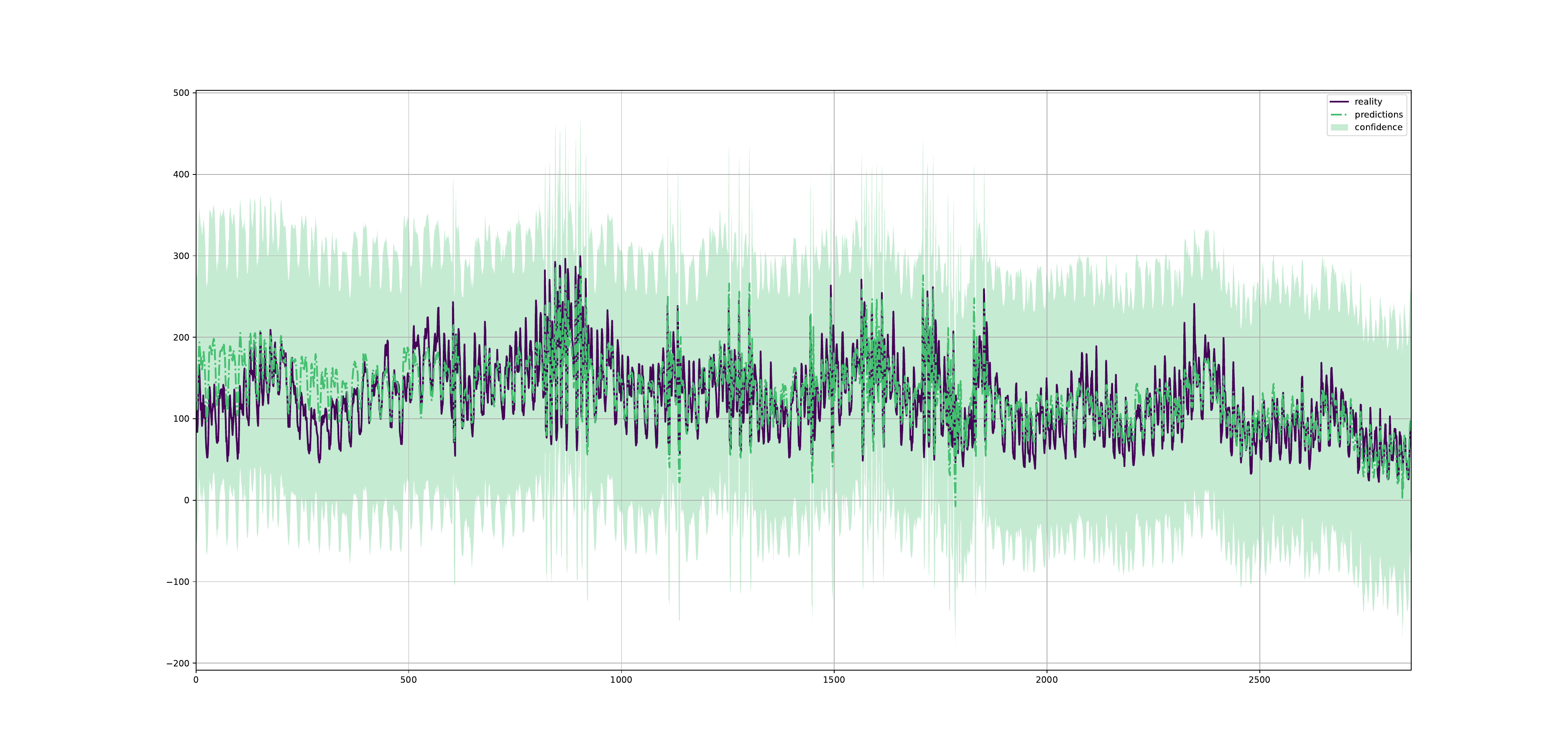}
        \caption{Substation A}
        %\label{fig:sub1}
    \end{subfigure}
    \hfill
    \begin{subfigure}{0.99\textwidth}
        \includegraphics[width=\textwidth]{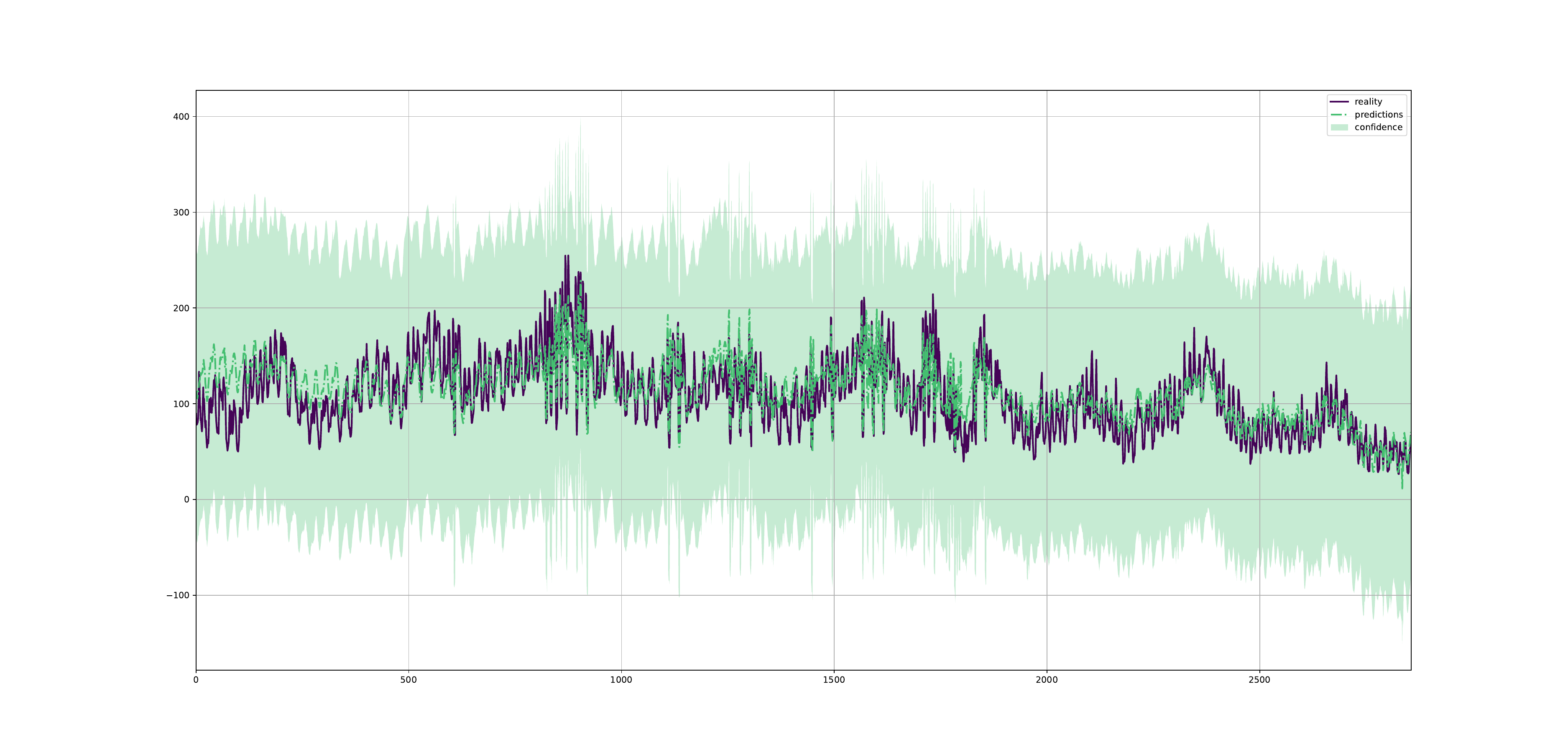}
        \caption{Substation B}
        %\label{fig:sub2}
    \end{subfigure}
    \hfill
        \begin{subfigure}{0.99\textwidth}
        \includegraphics[width=\textwidth]{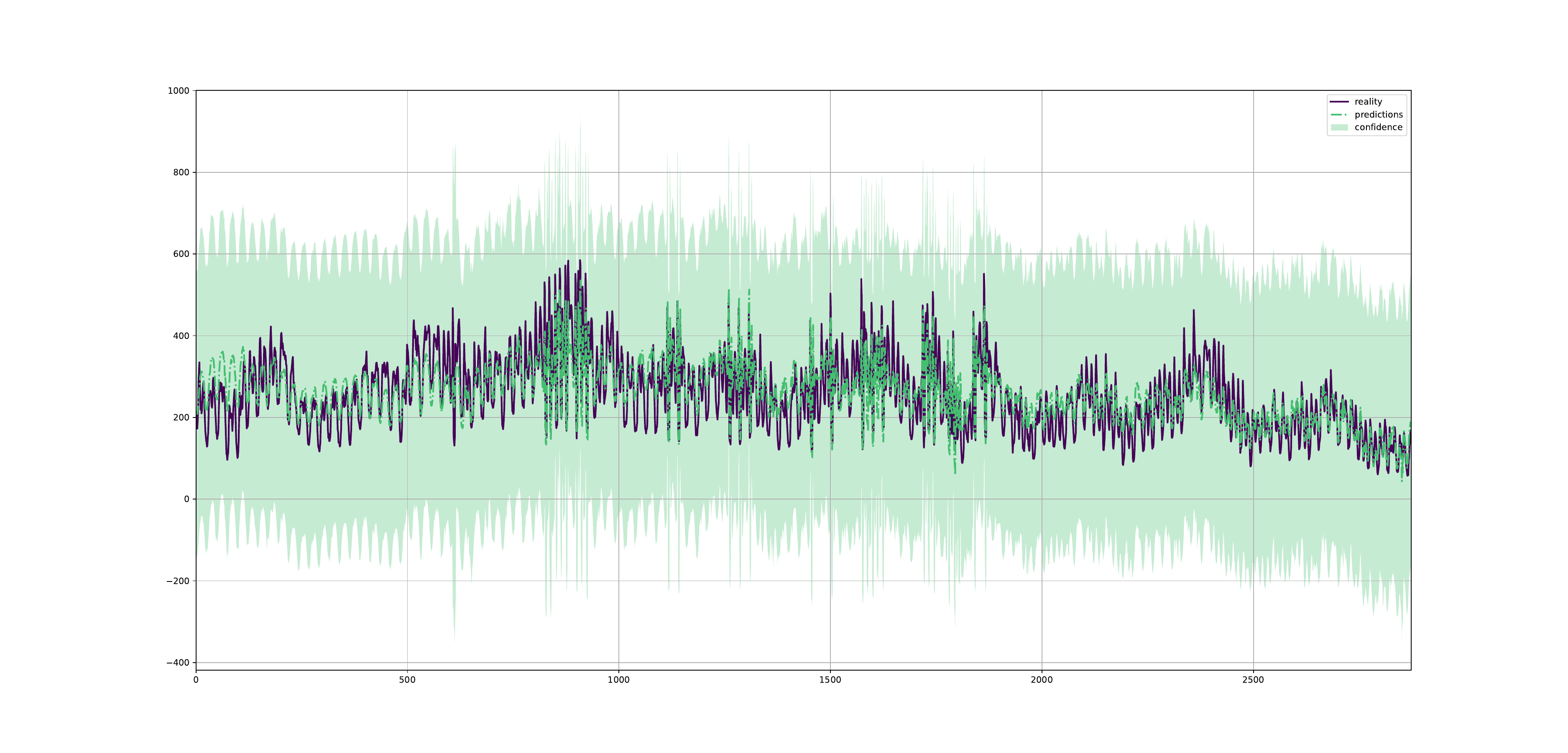}
        \caption{Substation C}
        %\label{fig:sub1}
    \end{subfigure}
    \caption{Test predictions of the benchmark at each substation}
    \label{fig:preds_conso}
\end{figure}

\begin{table}[h]
\renewcommand{\arraystretch}{1.2}
    \centering
    \caption{Absolute test errors of the best validation run for each substation}
    \begin{tabular}{c|ccc}
    \hline 

    \hline 
      \textbf{Substation} & \textbf{A} & \textbf{B} & \textbf{C}\\ \hline

      \textbf{MAE [kWh]} & 20.49103 & 17.90663 & 41.21746 \\

      \textbf{RMSE [kWh]} & 26.08270 & 22.39355  & 51.25044\\%680.30703 & 501.47115 & 2626.60727\\
    \hline 

    \hline 
    \end{tabular}
    \label{tab:bench}
\end{table}
As can be seen in both Figure \ref{fig:preds_conso} and Table \ref{tab:bench}, substation C is the hardest to predict for the Gaussian process. In general, the confidence of the model is also low (standard deviation is high). This hints at the complexity of the patterns.

\subsection{Acknowledgements:}
Special thanks to Odile Noël, Ahmed Abdellatif, Steve Boursiquot, and everyone from Hydro-Québec's open data initiative, especially Andrée-Anne Gauthier, and Robert Row. This work was partially funded by NSERC, FRQNT, Mitacs, and Hydro-Québec.

\end{document}